\pdfoutput=1

\documentclass[11pt,table,xcdraw]{article}

\usepackage{acl}
\usepackage{times}
\usepackage{latexsym}
\usepackage[T1]{fontenc}
\usepackage[utf8]{inputenc}
\usepackage{microtype}
\usepackage{inconsolata}
\usepackage{tabularx}
\usepackage{enumitem}
\usepackage{booktabs}
\usepackage{colortbl}
\usepackage{array}
\usepackage{longtable}
\usepackage{tcolorbox}
\usepackage[normalem]{ulem}
\usepackage{makecell}
\usepackage{graphicx}
\usepackage{longtable}
\usepackage{supertabular}

\definecolor{MasonGreen}{rgb}{0.12, 0.38, 0.22}

\newcommand{\ctrl}[0]{\texttt{c}}
\newcommand{\trtm}[0]{\texttt{t}}

\definecolor{control}{RGB}{186, 210, 232}
\definecolor{treatment}{RGB}{212, 171, 194}

\definecolor{ctrl}{RGB}{113, 154, 255}
\definecolor{trmt}{RGB}{179, 86, 140}
\definecolor{llms}{RGB}{0, 102, 51}
\definecolor{discuss}{RGB}{192, 0, 0}

\title{The LLM Effect: Are Humans Truly Using LLMs, or Are They Being Influenced By Them Instead?}

\author{
  Alexander S. Choi\textsuperscript{*$1$}, Syeda Sabrina Akter\textsuperscript{*$1$}, JP Singh\textsuperscript{$2$}, Antonios Anastasopoulos\textsuperscript{$1$,$3$} \\
  \textsuperscript{$1$}Department of Computer Science, George Mason University \\
  \textsuperscript{$2$}The Schar School of Policy and Government, George Mason University \\
  \textsuperscript{$3$}Archimedes AI Unit, Athena Research Center, Greece \\ 
  \texttt{\{achoi29,sakter6,jsingh19,antonis\}@gmu.edu}
}

\begin{document}
\maketitle

\begingroup
\renewcommand{\thefootnote}{*}
\footnotetext{Equal contribution.}
\endgroup

\begin{abstract}
Large Language Models (LLMs) have shown capabilities close to human performance in various analytical tasks, leading researchers to use them for time and labor-intensive analyses. However, their capability to handle highly specialized and open-ended tasks in domains like policy studies remains in question. This paper investigates the efficiency and accuracy of LLMs in specialized tasks through a structured user study focusing on Human-LLM partnership. The study, conducted in two stages—Topic Discovery and Topic Assignment—integrates LLMs with expert annotators to observe the impact of LLM suggestions on what is usually human-only analysis. Results indicate that LLM-generated topic lists have significant overlap with human generated topic lists, with minor hiccups in missing document-specific topics. However, LLM suggestions may significantly improve task completion speed, but at the same time introduce anchoring bias, potentially affecting the depth and nuance of the analysis, raising a critical question about the trade-off between increased efficiency and the risk of biased analysis.
\footnote{\url{https://github.com/achoigmu/llm_effect}}
\end{abstract}

\section{Introduction}
\label{sec:intro}

\begin{figure*}[t]
\centering
\includegraphics[width=\linewidth, keepaspectratio]{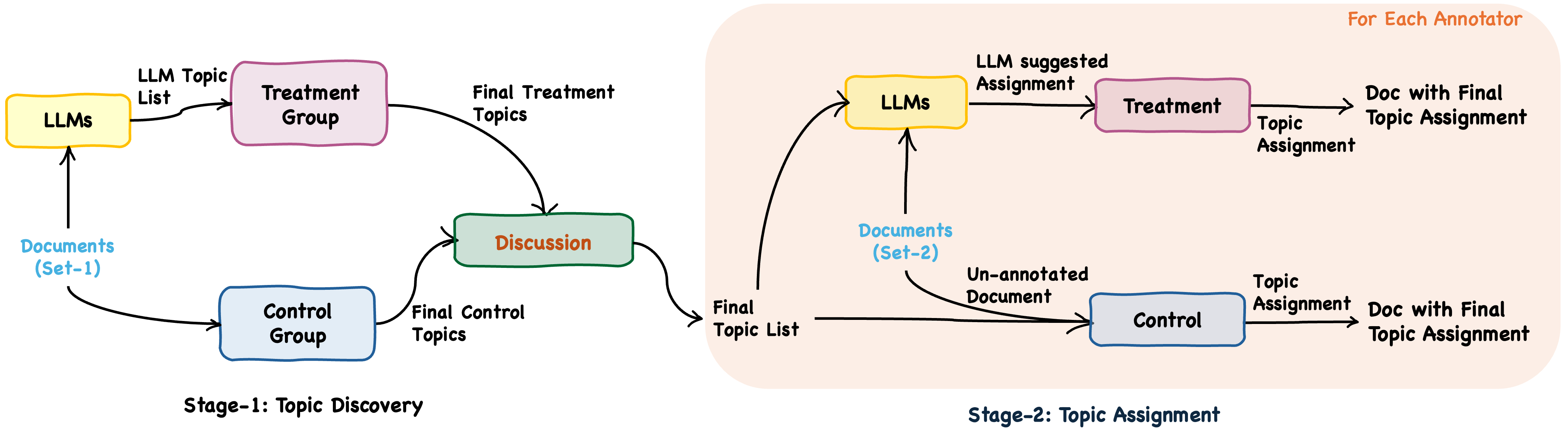}
\caption{An overview of the two stages of our user study. In both stages, we have the annotators read the documents and come up with a relevant topic list with (Treatment) and without (Control) the LLM suggestions. By the end of Stage 1, the annotators agree on a Final Topic List, which we use for our Topic Assignment stage. In Stage 2, all annotators conduct the task of assigning the topics to a separate set of documents with (Treatment) and without (Control) the LLM suggestions.}

\label{fig:pipeline}
\end{figure*}

Large language models (LLMs) like GPT-4~\cite{radford2019language},  LLaMA~\cite{touvron2023llama} etc., have recently dominated the research world by showcasing capabilities that are nearly equivalent to human performance in different analytical tasks. Researchers and organizations are increasingly using these models to conduct time-consuming analyses that were previously handled by human experts \cite{Rivera_escalation}. However, this raises a critical question: Are LLMs truly ready to undertake highly specialized tasks? Domains such as policy studies are inherently very complex and nuanced, requiring an adept proficiency that may extend beyond the current capabilities of LLMs. While these models can enhance efficiency and provide substantial support, their ability to match human expertise in specialized fields requires further scrutiny.

The advantages of using LLMs include increased efficiency, consistency in output, and the ability to handle large volumes of data quickly~\cite{brown2020}. On the other hand, using LLM suggestions as a helpful-guide for such open ended analysis has the potential to cause experts to rely heavily on the given suggestions, therefore, introducing anchoring bias~\cite{tversky1974judgment} for their task. 

To address these concerns, we designed a user study that integrates experts and LLMs in a highly structured way. Our key contributions are:

\begin{enumerate}[noitemsep,nolistsep]
    \item We evaluate the capability of a LLM at conducting open-ended, domain-specialized expert-level tasks and analysis by integrating it into a Topic Modeling study on ``AI Policies in India'' (see section~\ref{sec:data}).
    \item We investigate whether incorporating a LLM into an expert annotator's workflow increases their ability to complete their task more efficiently by comparing the time taken for topic assignment with and without LLM suggestions. 
    \item We examine the influence of LLMs on the decision-making processes of expert annotators to address the potential of cognitive biases introduced by LLM suggestions.
    \item To assess the level of trust and acceptance that expert annotators have for LLMs as an emerging technology, we conducted pre and post-study surveys.
\end{enumerate}

We chose Topic Modeling as our primary task for this study, as it is a standard method of analyzing larger documents for such human-led studies~\cite{brookes2019utility}. The study was conducted in two stages: Topic Discovery and Topic Assignment. In both stages, we integrated LLMs with human experts and observed how human-led analyses compared with and without LLM suggestions.

In summary, we found that with LLM suggestions experts performed the topic assignment task much faster than without them. However, a noticeable anchoring bias~\cite{tversky1974judgment} was observed in the analysis when experts worked with LLM suggestions. The bias introduced by LLM suggestions raises an important question: \textbf{Is the trade-off between the increased efficiency worth the potentially biased analysis?}

We also discovered that during the topic discovery stage, experts with LLM suggestions tended to keep the topics as they were, without making significant changes, even though the LLM suggestions were mostly very generalized and broad. Conversely, experts without LLM suggestions often came up with highly tailored topics specific to their given documents. This indicates that while LLMs are very effective for analyses requiring broad and generalized topics, they struggle with providing the depth needed for more nuanced tasks.

\section{Data and Tools}
\label{sec:data}

\paragraph{Data}
In 2022-2023, we conducted a series of eight interviews aimed at gaining unique and in-depth insights into the adaptation and impact of AI policy in India. These interviews were held between a policy studies expert and several prominent figures who play significant roles in shaping Indian AI policies\footnote{The interviews were chosen as part of a broader effort to analyze evolving AI policies worldwide, and because they offer content that had not been analyzed before or publicly available online.}. The discussions focused on understanding the values and priorities these influential individuals and their organizations (from private, government, and civil society sectors) hold concerning the development of AI policy. Initially, the interviews were recorded and subsequently transcribed using Automatic Speech Transcription technology~\cite{whisper} to ensure accuracy and facilitate analysis. Any sensitive information (such as names of individuals and organizations) were removed to preserve the anonymity of the interviewees.

\paragraph{AI Tools}

Topic Modeling~\cite{blei2012probabilistic} or analysis is the process of identifying patterns of word co-occurrences and using these patterns to group similar documents and infer topics within them. The most well-known algorithm for such Topic Modeling is Latent Dirichlet Allocation \cite[LDA;][]{BleiNJ03}, which examines word co-occurrences and groups documents accordingly. However, LDA often fails to capture the underlying context of documents, which is necessary for studying context-rich documents like those in our study. In addition, LDA yields a specific probability distribution over the words of the vocabulary that \textit{need to be interpreted} as a ``topic'', making it difficult to use from a practical perspective. Another approach is BERTopic~\cite{bertopic} that uses transformer models to understand the context within text and improve topic coherence. However, BERT-based models can also struggle with generating interpretable topic labels~\cite{DevlinCLT19}. In addition, the underlying model for BERTopic (BERT) has a very small context window, which leads to cumbersome heuristics needed for topic classification over longer documents. 

Instead of these techniques, we use a slightly modified version of TopicGPT~\cite{topicgpt}, a prompt-based framework leveraging GPT models to uncover latent topics in a text collection. It produces topics that align better with human categorizations compared to competing methods, while also generating interpretable topic labels and relevant definitions instead of ambiguous bags of words, making it a comprehensive tool for our Topic Modeling needs. The LLM model we use is \texttt{gpt-4-0125-preview} queried via the API. This GPT model has a context window of 128,000 tokens, which makes the feasibility of our study possible, given our 1-hour long interviews. %

\section{Study Design}
\label{sec:study design}

\begin{figure*}[t]
\centering
\includegraphics[width=\linewidth, keepaspectratio]{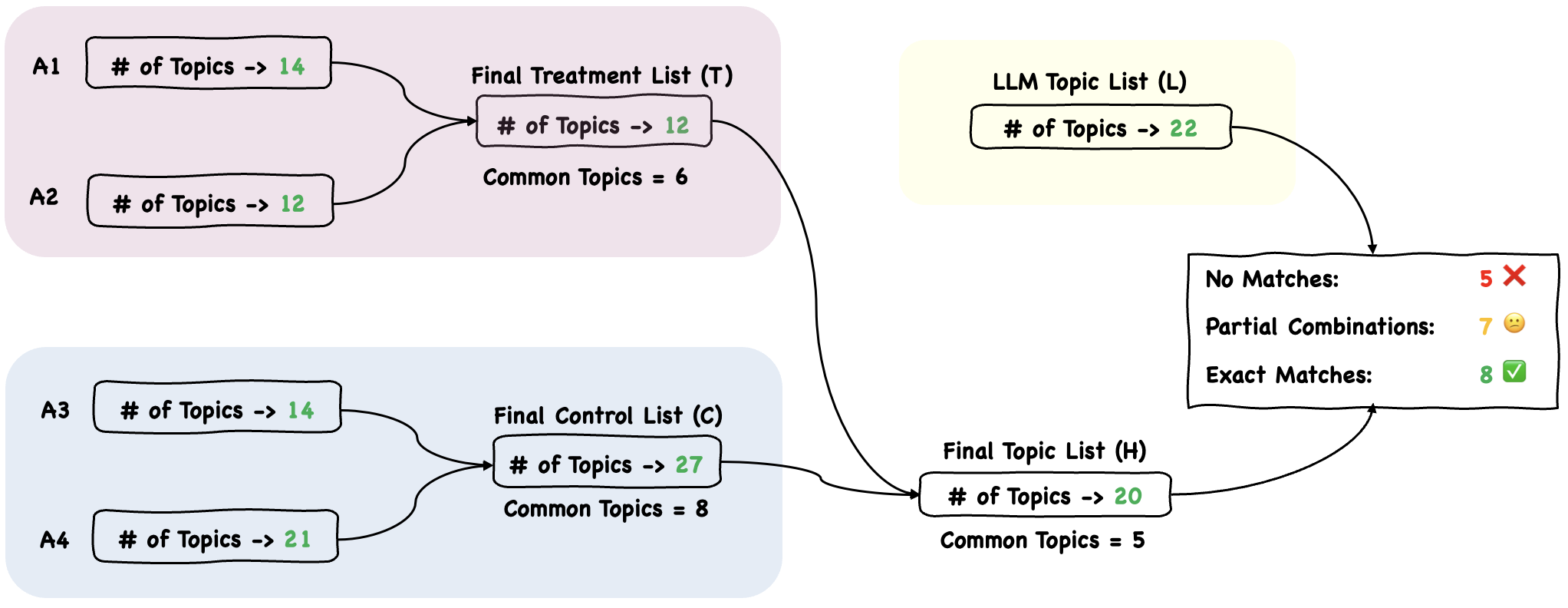}
\caption{The integration process of the topic lists from annotators in different settings for Stage 1. The Final Topic List (\texttt{H}) has some LLM topic overlaps due to the treatment team choosing to use many of the model generated topics and definitions. Most importantly, the LLM generated list doesn't cover 5 topics in any capacity that the control group deemed important.}
\label{fig:stage1_res}
\end{figure*}

Given the domain of the transcripts, we conducted the analysis focusing on topics relating to AI policy. We consulted four International Policy Experts to help annotate the transcripts with relevant topics. They were asked to ground their analysis within the realm of AI policy in India. The Annotators have extensive background knowledge in Policy Studies, with one being an expert on Indian Policies.

We conducted our study in two stages (see Figure~\ref{fig:pipeline}), each utilizing a research model with two settings. 

\begin{enumerate}[nolistsep,noitemsep,leftmargin=*]
    \item\textbf{Control Setting (\ctrl)}, the traditional setting that involves expert annotators conducting their analysis on the given documents without external suggestions from other tools or sources.
    \item \textbf{Treatment Setting (\trtm)}, a more custom setting in which we provide the LLM-generated suggestions to the expert annotators as a helpful guide.
\end{enumerate}

Note that, the annotators do not query the LLM directly. We designed a user-interface through Label Studio~\cite{LabelStudio} to help facilitate this study. The annotators accessed their documents through their own individual Label Studio interfaces. Specifically in the treatment setting, we provided the LLM suggestions as highlighted texts with corresponding label names. 

We instructed our experts to vocalize their thought process while conducting their analysis. This \textbf{Thinking Aloud Process}~\cite{johnson2013} during problem-solving requires annotators to continuously talk and verbalize whatever thoughts come to mind while doing the task. Unlike other verbal data gathering techniques, this method involves no interruptions or suggestive prompts. Annotators are encouraged to provide a concurrent account of their thoughts without interpreting or explaining their actions, focusing solely on the task at hand. Two research assistants served as scribes during the user study to document the experts' thought processes. This approach allows us to qualitatively study the strategies employed by the experts, providing insights into how they interpret and tackle the task of analyzing the documents.

We also developed pre- and post-analysis surveys to assess how familiar the expert annotators were with LLMs. The pre-survey aims to understand their initial assumptions regarding the use of LLMs versus conducting the analysis in the traditional way. With the post-survey, we wanted to gauge their reactions to the LLMs' suggestions and determine if they would be interested in using such technology in their future workflows.

\section{Stage 1: Topic Discovery}
\label{sec:stage1}

\paragraph{Methodology} For Stage 1, our goal was to have expert annotators build and curate a comprehensive topic list, generated over a set of documents, with and without the LLM suggestions. We also generated a similar topic list solely by an LLM - which was provided to the annotators in the treatment team - and have analyzed the similarity of both of the topic lists. Figure~\ref{fig:pipeline} shows the process of forming the final topic list which lays the foundation for subsequent analysis of Stage 2.

We allotted five hours for expert annotators to complete this stage of the study. We divided our four expert annotators into two teams: Annotators 1 (A1) and 2 (A2) conducted the topic discovery task under the treatment setting, while Annotators 3 (A3) and 4 (A4) completed the task under the control setting. The annotators were aware of each other's tasks, meaning the control annotators knew that the treatment group would receive LLM suggestions generated by the GPT model.

We applied TopicGPT~\cite{topicgpt} prompts to generate a LLM-provided topic list over the four Stage 1 documents. It is a two shot Topic Modeling prompt that generates a comprehensive topic list over a given document. We prompted the LLM four separate times for each of the 4 documents, and then we used a merging prompt to combine the four topics lists and remove any duplicate topics (See~\ref{sec:appendix2} and~\ref{sec:appendix3}). 
The final LLM generated topic list (\texttt{L}) (See Table~\ref{tab:final_topic_list}) contains 22 topics in total. We then used the topic assignment prompt (See~\ref{sec:appendix5}) to assign topic labels to each paragraph for the treatment team's documents which we then provided to the treatment group experts.

\paragraph{Control: Topic Discovery - Experts only}
The Annotators were instructed to read over their assigned document and generate a list of latent topics with corresponding definitions that exist within their document. They were also asked to highlight any sentence or paragraph they considered pertinent to a topic within their own generated topic list with the corresponding topic label.

\paragraph{Treatment: Topic Discovery - LLMs+Experts}
The experts in the treatment group were provided with the LLM-generated topic lists along with LLM annotated transcripts to help guide their topic generation. The control group received no LLM aid in completing the same task. Annotators did not interact with each other in this step.

\paragraph{Combining Control and Treatment}
After experts completed their tasks individually, they were asked to discuss and develop a combined topic list for their settings. A1 and A2 decided on the final treatment list (\texttt{T}), while A3 and A4 finalized the control list (\texttt{C}). Finally, all four annotators reviewed both the control and treatment lists, discussing their processes, documents, and definitions. During this back-and-forth, the experts made a variety of decisions, determining if a concept or 'topic' was too generalized and needed to be broken down into multiple topics, if two or more concepts should be combined into one, or if certain topics should be renamed, all while developing final definitions for each topic.
This was a holistic process, rather than simply combining or adding the two lists up. Through these discussions, they created the final golden human curated Stage 1 topic list. We refer it as the \textbf{Final Topic List (\texttt{H})} from here onwards.

\begin{table}[t]
    \begin{tabularx}{\linewidth}{cXc}
    \toprule
     & Comparing \texttt{H} and \texttt{L} & \# of Topics \\
    \midrule
    1 & Exact matches between \texttt{H} \& \texttt{L} & 8 \\
    2 & Present in \texttt{H}, but not \texttt{L} & 5 \\
    3 & Single \texttt{H} topic encompassing two or more \texttt{L} topics & 5 \\
    4 & \texttt{H} topics split from a broader, more generalized \texttt{L} topic & 2 \\
    \midrule
     & \textbf{Total} & \textbf{20} \\
    \bottomrule
    \end{tabularx}
    \caption{The comparison of the LLM topic list (\texttt{L}) with respect to the Final Topic List (\texttt{H}) show that there are a very small number of topics that the model has failed to cover in its overall topic generation task.}
    \label{tab:topic_matching}
\end{table}

\paragraph{Results and Analysis}
\label{subsec:stage1_results}

By the end of Stage 1, we obtained two topic lists: one from the control group (\texttt{C}, no LLMs involved) and one from the treatment group (\texttt{T}, with LLM aid). In addition, we also have the Final Topic List(\texttt{H}), curated by the annotators based off of the two aforementioned lists. Figure~\ref{fig:stage1_res} shows the process of how these lists were developed and integrated to form the final topic list (\texttt{H}).
The results reveal a broad spectrum of topics identified through both control and treatment settings. The control lists identified 14 and 21 topics individually. When consolidated, the annotators unified their 8 common topics and curated the Final Control List (\texttt{C}) comprising of 27 topics. 

The LLM generated topic List (\texttt{L}) identified 22 topics over the same set of documents given to the experts for Stage 1. In the treatment setting, annotators identified 14 and 12 topics individually, most of which aligned with the LLM-generated topic list (\texttt{L}). This alignment happened because the treatment group, having received LLM suggestions, tended to rely more on them than coming up with topics on their own. Most of their "editing" work was focused on grouping or removing LLM-suggested topics instead of coming up with new ones. The Final Treatment List (\texttt{T}) resulted in 12 topics, with 6 topics shared initially between the annotators. The combined Final Topic List (\texttt{H}), included 20 topics, with 5 topics common to both settings. 

\begin{table}[t]
    \centering
    \begin{tabularx}{0.97\linewidth}{@{}r@{ }Xcc@{}}
    \toprule
    & Missing Topics            & Stage 1 & Stage 2  \\
    \midrule
    1 & Civil Society Advocacy    & 16.4\%              & 5.1\% \\
    2 & Transportation            & 1.8\%               & 2.3\% \\ 
    3 & Policy Institutions       & 5.5\%               & 6.7\% \\
    4 & Policing \& Surveillance  & 6.0\%               & 7.3\% \\
    5 & Academia                  & 3.6\%               & 2.3\% \\            
    \midrule
\multicolumn{2}{c}{\textbf{Average Topic Coverage}} & \textbf{6.7\%} & \textbf{4.7\%} \\
    \bottomrule
    \end{tabularx}
    \caption{Topic assignment coverage percentage of the Missing Topics in the two sets of documents. Note that, for Stage~2 we use the results of the control setting.}
    \label{tab:missing topics}
\end{table}

We wanted to evaluate how well the LLMs captured the topics of the given documents compared to the expert annotators. For this, we compared both sets of topics generated in Stage 1. We consider the Final Topic List (\texttt{H}) as the gold standard as it was curated by all experts following considerable discussion among them. We found that the LLM-generated topics (\texttt{L}) fall into four different categories (see Table~\ref{tab:topic_matching}) with respect to the Final Topic List (\texttt{H}). Among the 20 \texttt{H} topics, 15 were covered by the LLM in \texttt{L} either directly or through overlap with multiple combinations of topics. However, there were 5 \texttt{H} topics that were not covered by the LLM in \texttt{L} in any form. The `missing' topics are listed in Table~\ref{tab:missing topics}.

To understand the significance of the topics labeled as `missing' in Table~\ref{tab:missing topics}, which refers to topics that were underrepresented or not covered by the LLMs in our analysis, we examined their assignment in the documents of Stage 1 and Stage 2 control settings, both of which were done by the expert annotators. We analyzed how frequently these 5 missing topics appeared in the documents. We found that these topics had a rather low assignment percentage coverage (see Table~\ref{tab:missing topics}).

Our analysis shows that while LLMs are effective in capturing a majority of the topics identified by experts, they still lack the ability to uncover possibly critical nuances latent within documents. The 5 topics in \texttt{H} that remained completely undetected to the LLMs tended to have low total prevalence counts within the documents as a whole (see Table~\ref{tab:missing topics}), suggesting that these topics might be subtle or context-specific, and require human expertise for identification. This highlights \textbf{the importance of integrating human insights with LLM capabilities to ensure a comprehensive and nuanced understanding of the subject matter}.

It is important to mention that the topics generated by the LLMs were more generalized and did not have clear distinctions from one another. It often happened that a few topics in \texttt{L} had overlapping definitions. In contrast, all of the human-generated topic lists (\texttt{C} and \texttt{H}) were more distinct and clearly separated by their definitions.

\section{Stage 2: Topic Assignment}
\label{sec:stage2}

\begin{table}[t]
    \centering
    \begin{tabular}{c|cccc}
    \toprule
        & \multicolumn{4}{c}{Annotator} \\
        & I & II & III & IV \\
    \midrule
        D5 &  & \cellcolor{control}\ctrl & & \cellcolor{treatment}\trtm  \\
        D6 & \cellcolor{control}\ctrl & & \cellcolor{treatment}\trtm & \\
        D7 & & \cellcolor{treatment}\trtm & & \cellcolor{control}\ctrl \\
        D8 & \cellcolor{treatment}\trtm & & \cellcolor{control}\ctrl & \\
    \bottomrule
    \end{tabular}
    \caption{For Stage 2, each expert gets two documents to annotate; one for their control setting and the other for their treatment setting. With this combination, we get each document annotated at least once in both settings.}
    \label{tab:assignment}
\end{table}

\paragraph{Methodology}
\label{subsec:stage2_methodology}

In Stage 2, we studied how the topic assignments vary for annotators in both control and treatment settings. For this stage, we used 4 documents, different from those used in Stage~1. Each annotator received 2 documents, and they were instructed to work on these individually sans discussion with other annotators. Annotators were also instructed to conduct topic assignments on the two documents in two different settings: one as control and the other as treatment (see Table~\ref{tab:assignment}). We used a Latin squares study design~\cite{montgomery2017design} methodology in order to abstract away potential annotator-specific variability. 

To accomplish our research goal of measuring the LLM accuracy of topic assignments, we instructed both expert annotators and the LLM to assign topics on a per-paragraph basis. This would allow for a granular enough approach to collect a meaningful amount of data points per document, while ensuring enough context for both experts and the LLM to comfortably make topic assignment decisions. On average, our Stage 2 transcripts contained 44 paragraphs. 

For the treatment setting, we generated topic assignments over the same set of transcripts by prompting the LLM with a topic assignment prompt (see Appendix~\ref{sec:appendix5}). The model was provided with the Final Topic List (\texttt{H}) along with the transcripts \textbf{at a per paragraph level.} Multiple topic assignments per paragraph are allowed.

\begin{table}[t]
    \centering
    \begin{tabular}{ccc|cc}
    \toprule 
    \multicolumn{5}{c}{LLM Precision \& Recall measured against} \\
    \midrule
    & \multicolumn{2}{c|}{Control} & \multicolumn{2}{c}{Treatment} \\
    doc & precision & recall & precision & recall \\
    \midrule
    D5  & 31.4            & 56.3            & 84.9        & 83.9  \\
    D6  & 48.1            & 62.6            & 68.2        & 72.7  \\
    D7  & 27.9            & 51.5            & 61.5        & 88.2  \\
    D8  & 68.4            & 60.5            & 71.1        & 73.0  \\
    \cmidrule{1-5}
    Avg & \textbf{44.0} & \textbf{57.7} & \textbf{71.4} & \textbf{79.5} \\
    \bottomrule
    \end{tabular}
    \caption{For each transcript used in Stage 2, the precision and recall percentages of the LLM annotations over these transcripts when measured against the annotations of experts either acting under the control or treatment setting. Also, the averages of these LLM precision and recall percentages,}
    \label{tab:precision_recall}
\end{table}

\paragraph{Control: Topic Assignment - Expert Only}
For the control setting, annotators received a transcript and the Final Topic List (\texttt{H}) with definitions (see Table~\ref{tab:final_topic_list}). Annotators were to assign topics to the transcript with the possibility of multiple topics per paragraph. 

\paragraph{Treatment: Topic Assignment - Experts+LLM} 
In the treatment setting, we provided the LLM-generated assignments to the experts to annotate each document at a paragraph level with topics from the same topic list as LLMs, allowing multiple topics per paragraph. Annotators received the LLM annotations as suggestions and were tasked with cross-checking and, if necessary, correcting the assignments.

\paragraph{Experimental Setting} The annotators who were in the control team in Stage 1 were asked to complete the treatment task first and then the control task. The treatment team of Stage 1 was asked to do the opposite. Additionally, we tracked the time taken to complete each stage for each document. After all annotators completed all Stage 2 tasks, we collected the annotated documents and summarized the results. 

We created a 21 element vector for each paragraph within an annotated document. 20 of the elements correspond to the list of 20 topics in the final topic list agreed upon by all experts at the end of Stage 1; one element represented ``None'', indicating none of the 20 topics corresponded to that paragraph. Each element in a vector represents either the existence or absence of a topic within that paragraph. Both the expert annotators and the LLM usually assigned between one to three topics per paragraph. This data representation allowed us to perform various statistical analyses on the transcripts.

\begin{table}[!t]
    \centering
    \begin{tabular}{ccc}
    \toprule
    \multicolumn{3}{c}{Average Annotation Speed (words/min)} \\
    \midrule
    Control & Treatment &  Increase (\%) \\
    96.4 & 225.0 & 133.5\% \\
    \bottomrule
    \end{tabular}
    \caption{Comparison of average annotation speeds between control \& treatment settings, measured in words per minute.}
    \label{tab:avg annotation speed}
\end{table}

\paragraph{Results and Analysis}
\label{subsec:stage2_results}

Upon inspection of our results, we find both promising data, but also alarming trends. When measuring LLM topic label accuracy against the \textit{control} annotations, \textbf{the average precision and recall were 44.0\% and 57.7\%, respectively} (see Table~\ref{tab:precision_recall}). These are encouraging numbers, considering the incredibly open-ended nature of the task. 

\begin{table*}[t]
\centering
\begin{tabular}{l|cccc|cccc}
\toprule
 & \multicolumn{4}{c}{Annotator Agreement with LLM} & \multicolumn{4}{c}{Annotation Speed (words/min)} \\
 & A1 & A2 & A3 & A4 & A1 & A2 & A3 & A4 \\
\midrule
D5 &  & \cellcolor{control}36.6\% &  & \cellcolor{treatment}84.4\% &  & \cellcolor{control}92.31 &  & \cellcolor{treatment}207.7 \\
D6 & \cellcolor{control}50.2\% &  & \cellcolor{treatment}62.2\% &  & \cellcolor{control}110 &  & \cellcolor{treatment}330 & \\
D7 &  & \cellcolor{treatment}70.7\% &  & \cellcolor{control}29.0\% &  & \cellcolor{treatment}214.7 &  & \cellcolor{control}250.5 \\
D8 & \cellcolor{treatment}68.9\% &  & \cellcolor{control}59.6\% &  & \cellcolor{treatment}130.15 &  & \cellcolor{control}86.76 & \\
\bottomrule
\end{tabular}
\caption{Topic assignment Stage 2 results. In the left table, the percentages represent the Cohen's $\kappa$ \cite{cohen1960coefficient} level of agreement between the expert and the LLM within different settings. The right table shows annotation speed (words per minute) of each expert within each document and setting. The control setting is highlighted in blue, while the treatment setting is highlighted in pink. A noteworthy trend - when annotators had LLM suggestions they tended to heavily agree with the LLM, and in correlation with this heavy LLM agreement, annotation speed tended to increase significantly.}
\label{tab:llm_agreements}
\end{table*}

We also find that annotation speed improves markedly with LLM suggestions. On average, the annotators operated at a pace of 96.4 words per minute in the control setting.\footnote{It should be noted that A4 was interrupted throughout the completion of their Stage 2 tasks. It took them around 30 minutes to complete annotations for both the control and treatment. We decided to exclude their annotation speed from our final assessment.} Conversely, in the treatment setting, the annotators operated at a pace of 225.0 words per minute on average. \textbf{This difference represents an annotation efficiency increase of 133.5\%} (see Table~\ref{tab:avg annotation speed}).

However, disconcerting trends arise through the analysis as well. In contrast to LLM accuracy measured against the control, the LLM's performance against the \textit{treatment} annotations showed a precision of 71.4\% and recall of 79.5\%, significantly higher than the control annotations. We go a step further and employ Cohen's $\kappa$~\cite{cohen1960coefficient} coefficient to analyze similarities between annotations of the same document (see Table~\ref{tab:llm_agreements}). When annotators act under the control setting, the similarity of their annotated transcripts compared with the LLM's annotated transcripts averages to 43.9\%. Yet, when the annotators act under the treatment setting, their agreement with the LLM, on average, rises to 71.5\%, indicating the annotators and LLM aligned heavily. 

This substantial discrepancy leads us to evaluate the difference between the two settings. Thus we employ statistical significance tests to investigate the existence of a non-random difference between the two distributions. Each expert annotated, on average, 44 paragraphs within each setting, leading to 176 annotated data points per setting. We conduct a paired sample $t$-test over the paragraph level Cohen's $\kappa$ numbers.\footnote{This test is appropriate because each of our four annotators acted as both control and treatment.} Running the paired $t$-test, we get a $p$-value = 1.087e-14. Thus, we can (safely reject the null hypothesis that the two samples were drawn from the same distribution and) conclude the existence of a statistically significant non-random difference between the control and treatment annotation agreements.

One possible interpretation of these results is that the LLMs provide fairly accurate Topic Modeling outputs, according to the annotators. 
However, this does not explain the significant reduction in alignment when the annotators act as control. To explain this, we have proven statistically that there exists a difference between the two settings that is non-random, and as a result of our study design, the only variable that has changed is the introduction of LLM suggestions. If this is the only variable that has changed, then the presense of LLM suggestions themselves must be the cause for such high treatment-LLM alignment. Therefore, \textbf{we conclude that when an expert annotator receives LLM suggestions to aid their individual decision making process, they tend to become anchored to and biased by these LLM outputs}.

\section{Discussion}
\label{sec:discussion}
It is apparent there are multiple factors at play when it comes to utilizing LLMs for open-ended tasks such as Topic Modeling. In terms of promising impact presented by LLMs, we put the difficulty of this task fully into perspective. Given a document with dozens of paragraphs, the LLM must decide on a label or combination of labels out of a possible 20 choices to assign to each paragraph. \textbf{When we measure the accuracy of these LLM label assignments against 4 independent expertly annotated control documents, we get an average recall of 57.7\%} (see Table~\ref{tab:precision_recall}). Given the nature of the task, we consider this high from a research perspective, while also recognizing that from a practical implementation perspective, it may only be considered adequate. So, of course, we would like overall accuracy to improve. We leave this for future work.

Coupled with reasonable accuracy, we observe substantial increases in workflow efficiency. \textbf{We recorded a 133.5\% words per minute annotation speed increase when annotators utilized LLM suggestions. }This presents one possibility of massive reductions in labor intensive and time consuming workloads. 

However, if the goal is to obtain gains in workflow efficiency, this will come at significant cost. As mentioned earlier in Section 5, 
we see a significant difference between control and treatment annotation decisions (see Table~\ref{tab:llm_agreements}). Whether we examine annotator-LLM agreement over a particular document or over a particular annotator, the trend toward LLM bias remains consistent. For example, with regard to document 5, the agreement between the control annotations and the LLM annotations is 36.6\% while the the agreement between the treatment and LLM is 84.4\%. Additionally, if for example, we look at annotator 2, their agreement with the LLM when acting as control is 36.6\% while their agreement when acting as the treatment, is 70.7\%. \textbf{In \textit{every single instance}, the treatment agreement is higher than its control counterpart.} We find the implications of this trend worrisome. 

Additionally, as shown in our Stage 1 results, five topics that human annotators decided to add to the final topic list were not generated by the LLM. These five topics reflected the effort of a nuanced examination of the transcripts provided to the expert annotators. For example, "Policing and Surveillance" was not captured by the LLM (see Table~\ref{tab:llm_topic_list}). During the final discussion phase of Stage 1, scribes noted that annotators adamantly defended the inclusion of this topic in their final topic list (see Table~\ref{tab:final_topic_list}), even though the topic covered a relatively small portion of the transcripts (see Table~\ref{tab:missing topics}). Another point of contention was the LLM's decision to output "Gender Studies" as a topic label (see Table~\ref{tab:llm_topic_list}). Without capability of sensitivity or nuance, the LLM assigned "Gender Studies" to multiple topics that were regarded as topics that should more appropriately be labelled as "Gender Issues." \textbf{Thus, our findings suggest SOTA LLMs are able to reveal broad and generalized topics from lengthy domain specialized documents, however they still lack the ability to capture low prevalence high importance concepts.}

\paragraph{Survey Result}
We conducted pre- and post-study analysis surveys to evaluate the change between the expert annotators' initial perceptions and their actual experiences utilizing LLM suggestions and how this experience influenced their trust and reliance on LLM technology for complex tasks. The results can be found in Appendix~\ref{sec:survey}.

In the pre-analysis survey, all experts had prior experience with LLMs and expressed preferences for using them in their workflows. However, trust in the technology remained skeptical, with 50\% expressing neutral trust levels and concerns about reliability, accuracy, and the potential for LLMs to limit creativity or introduce bias. Confusion over LLM outputs was a moderate concern, with 50\% expecting them to be slightly confusing.

In the post-analysis survey, preferences for LLM recommendations remained strong at 100\%. Trust and reliability ratings showed slight improvements, with fewer experts finding the outputs confusing and an increase in perceived accuracy. However, concerns about biases and over-reliance on LLM suggestions persisted, indicating that while LLMs were appreciated for their efficiency, human oversight remained critical.

The feedback also showed some changes in attitudes. Several experts who were initially skeptical about biases found the LLM recommendations actually supported their work by improving task efficiency or prompting deeper thinking. However, the need for critical evaluation of LLM suggestions was a common theme, with annotators emphasizing the importance of balancing usage with expert judgment.

\paragraph{Think Aloud Process Findings}

During the think aloud process, experts displayed varied approaches with some differences between the control and treatment group. In the control group, annotators struggled with deciding between broad vs fine-grained labels, especially when topics overlapped or when content was nuanced. One expert preferred to work sentence by sentence, applying specific labels, while another read the entire document first to grasp context, before labeling. A common challenge was determining how much detail to include in the labels, with one opting for more value-neutral terms while the other focused on capturing opinions or sentiments expressed in the text.

In the treatment group, annotators initially questioned the generalized labels provided by the LLM. However, over time, they grew more comfortable with the LLM’s suggestions, finding that they aligned with a significant portion of their own thoughts. Despite this, there were still concerns about the LLM being too reactive to specific words, producing overly broad labels. Both annotators in this group appreciated the efficiency of the LLM but emphasized the importance of refining its output manually to ensure accuracy.

\section{Related Work}
\label{sec:related_works}
\paragraph{Topic Modeling}
The motivation underpinning Topic Modeling is the notion that concepts or latent "Topics" exist within a document. As mentioned in section~\ref{sec:data}, LDA is a popular machine learning methodology for Topic Modeling. 
However, implementing LDA models for real world applications has proven impractical, because of their inherent lack of interpretability~\cite{Gao_topics_applied_2024, Poursabzi2021, ross_interpretable}. 
As language models become more powerful and capable, some researchers have begun to develop ways to utilize these AI tools to approach the broad problem of latent topic discovery and assignment. TopicGPT~\cite{topicgpt} introduced a LLM prompting framework which utilizes the power of pretrained GPT models. Both CollabCoder~\cite{gaoCollabCoder} and SenseMate~\cite{OverneySenseMate} propose a human-in-the-loop coding pipeline that is geared towards novice annotators for simple and short coding tasks. CollabCoder goes a bit further and also suggests group work as part of its pipeline. While these studies utilized older language models, the potential efficiency gains observed in these studies help reinforce the findings of our own study.  

\paragraph{Human-LLM Partnership}
Much optimism surrounds the conversation regarding Human-LLM partnerships and many recent user studies have explored the benefits of integrating LLMs into human workflows~\cite{vats2024_survey_human_ai_teaming}. Microsoft has also published two technical reports regarding employee experiences using "Copilot", their GPT-powered AI assistant. In one study they found employees "read 11\% fewer individual emails and spent 4\% less time interacting with them"~\cite{cambon2023early, jaffe2024generative}. 

\paragraph{Anchoring Bias}
Anchoring bias is a phenomenon of human behavior in which, during the decision making process, a human is introduced to an initial piece of information, and future decisions are heavily influenced by the "anchor" this initial piece of information establishes~\cite{tversky1974judgment}. The theory of anchoring bias has been around for many decades and has been observed in many contexts and situations~\cite{FURNHAM201135}. In one study with law enforcement agents (LEAs) and mapping algorithms,~\citet{haque2024} found that LEAs became easily anchored to the initial algorithmic mapping output. Even after numerous suggestions from the researchers to consider options beyond the initial output, LEAs were still anchored to the first piece of information they saw. In another study,~\citet{enough} found that in legal courtroom settings, judges were found to be influenced by some initial information, and passed sentencing judgements while being anchored to that initial information. In a third study,~\citet{MUSSWEILER_subliminal} used a more general knowledge non-expert setting and found subliminal anchoring bias occurring.

Connecting the theory of anchoring bias with LLMs, we next mention the well-researched areas describing many of the potential dangers regarding LLM usage and outputs. LLM hallucinations have been well documented~\cite{ji_hallucination}, and their inherent bias regarding culture, gender, race, etc. has been heavily studied and confirmed~\cite{mukherjee_global}. \citet{resnik2024} states "For all their power and potential, large language models (LLMs) come with a big catch: they contain harmful biases that can emerge unpredictably in their behavior." Along with dangerous content, they have also been found to be overconfident and persuasive~\cite{jakesch2023, hancock2020}. Thus, toxicity, hallucinations, and persuasiveness are a potently dangerous combination. As \citet{jakesch2023} state "With the emergence of large language models that produce human-like language, interactions with technology may influence not only behavior but also opinions: when language models produce some views more often than others, they may persuade their users." So, the possibility of toxic and incorrect content combined with persuasive execution of language output can potentially lead to a pernicious influence on end users that is both worrisome and unpredictable.

\section{Conclusion and Future Work}
\label{sec:conclusion}
Our study highlights the trade-offs of integrating LLMs into expert Topic Modeling workflows. LLMs have made incredible strides in open ended tasks such as discovering and assigning generalized topics over documents. However, as the capabilities of LLMs continue to improve, safeguards against LLM anchoring bias must also be researched and implemented. We are excited for future research that further investigates both the use of LLMs for such tasks, while also investigating strategies that can mitigate this cognitive bias.

\section*{Limitations}
While our study demonstrates the potential of LLMs in enhancing the efficiency of expert Topic Modeling, it is limited by the scope of the data, focusing solely on AI policy in India. This may affect the applicability of our findings to other domains and geographic contexts. The study also requires computational resources in the form of OpenAI API credits, making it less accessible for smaller independent research teams. Over the course of this research project, we spent approximately \$100 testing and querying various GPT models. Another limitation is that our results are based on a relatively small number of documents and annotators, which may limit the statistical robustness of our conclusions. Finally, it would have been interesting to query other LLMs for comparison, however, at the time of our study, no other LLM came close to achieving the context window of 128,000 tokens. Due to the length of our documents and the difficulty finding annotators, from a practical feasibility perspective, no other LLM options existed. Also, while longer interviews allowed for the collection of many data points per transcript, it also requires more time for annotators to work through. We hoped to be able to cover more documents in Stage~1, however time was a limitation.

\section*{Ethics Statement}
Our research does not involve any practices that could raise ethical concerns, and we have completed the responsible NLP research checklist to affirm our adherence to these standards. 
This study was exempted by the appropriate ethics board. Thus, we do not anticipate any ethical issues arising from our work, and are prepared to address any inquiries from the Ethics Advisory Committee should the need arise.

\section*{Acknowledgements}
We are thankful to the reviewers and meta-reviewer for their constructive feedback. We are also thankful to our four policy experts, with whom this study would not have been possible. 

This research is supported by a grant from the Minerva Research Initiative of the Department of Defense (Award No: FA9550-22-1-0171).
It has also benefited from resources provided through the Microsoft Accelerate Foundation Models Research (AFMR) grant program.
Antonios Anastasopoulos is additionally generously supported by the National Science Foundation under grant IIS-2327143.
This work was partially supported by resources provided by the Office of Research Computing at George Mason University (URL: \url{https://orc.gmu.edu}) and funded in part by grants from the National Science Foundation (Award Number 2018631).

\bibliography{custom}

\pagebreak

\appendix

\onecolumn
\section{Example Topics}
\label{sec:appendix1}

The following are the topics that were provided to the expert annotators as an example in Stage 1. 

\begin{tcolorbox}
    \footnotesize
    \texttt{\textbf{Startup Ecosystem Development}: Focuses on the support and growth of startups through policies, incubation programs, and partnerships. This includes fostering innovation, providing resources for startups, and creating an environment conducive to entrepreneurial success.\\
    \\
     \textbf{Data Governance and Privacy}: Addresses the management, sharing, and protection of data in the digital age. This includes the development of policies and frameworks to ensure data privacy, security, and ethical use of data.}
\end{tcolorbox}

\section{Study Script}

\begin{tcolorbox}
    \footnotesize
    \texttt{Hello. My name is ----, this is ---- and ----. We are currently doing research on how we can integrate LLM assistants as part of experts' long document analysis workflow. Thank you for taking time out of your schedule to contribute to this study. During the course of this study, we may ask you questions about your experiences. We do not mean to insult or offend you, but instead to try to make you think deeply about why you do what you do. Try not to take anything personal and answer as best you can; there are no right answers.\\
    \\
     We ask that through the study, you voice your thoughts about the task you are performing and the data we put in front of you. ---- and ---- will monitor the interactions and take notes for posterity.\\
     \\
     \textbf{The Thinking Aloud Process:} The participants are asked to talk aloud, while solving a problem and this request is repeated if necessary during the problem-solving process thus encouraging the study participants to tell what they are thinking.\\
     \\
    Thinking aloud during problem-solving means that the participant keeps on talking, speaks out loud whatever thoughts come to mind, while performing the task at hand. \\
    \\
    Unlike the other techniques for gathering verbal data, there are no interruptions or suggestive prompts or questions as the participant is encouraged to give a concurrent account of their thoughts and to avoid interpretation or explanation of what they are doing, they just have to concentrate on the task. \\
    \textbf{This seems harder than it is.}\\
    It becomes a routine in a few minutes. Because almost all of the subject’s conscious effort is aimed at solving the problem, there is no room left for reflecting on what they are doing. \\
\\
    \textbf{Notice that these interviews are confidential, and we ask for your discretion with regards to the topics discussed here; because of our IRB protocol, the content of these interviews cannot be shared outside of this research exercise. }\\
    \\
    \textbf{Defining the task:} Our goal is to analyze documents. In particular we will perform an analysis over 8 interviews using ``topic analysis". \\
    \\
    Here, we are interested on topics relating to AI policy. These interviews give us in-depth insights into how AI policy is formulated, and we aim to determine the values and priorities that go into developing AI policy. \\
    \\
    An example of such a topic could be:
    \textbf{[See Appendix~\ref{sec:appendix1} for example topics]}\\
    \\
    We will first assign you in two teams:
    \begin{enumerate}[nolistsep,noitemsep,leftmargin=*]
        \item Team 1 [control]: ----, ----
        \item Team 2 [treatment]: ----, ----\\
    \end{enumerate}
    Each team will receive four interviews, and each annotator will be able to read two of them. In this stage, we are interested in ``topic discovery". Ultimately, we want a list of ``topics" as they show up in your documents.  After working on your two documents individually, you will have to get together with your team member to produce a final list of topics.\\
}
\end{tcolorbox}

\begin{tcolorbox}
    \footnotesize
    \texttt{And then, both groups will get together to create a final-final list of topics along with their definitions. This will conclude the first part of the study, and we will break for lunch. In the second part of the study, we will explore some new documents, and assign their sections with the pre-decided topic labels.\\
    \\
    \textbf{Interface:}
    We will use \texttt{labelstudio} for both annotation stages.
    \begin{itemize}[nolistsep,noitemsep,leftmargin=*]
        \item Please use this link to sign up: ----
        \item Navigate to the ``Sample Interview Topic Annotation" project, so we can familiarize ourselves with the annotation interface, and then we'll dive in.
    \end{itemize}}

\end{tcolorbox}

\vspace{-120pt}

\section{Topics Generation Prompt}
\vspace{-80pt}

\begin{tcolorbox}
    \footnotesize
\texttt{You will receive a document and a set of top-level topics from a topic hierarchy. Your task is to identify generalizable topics within the document that can act as top-level topics in the hierarchy. If any relevant topics are missing from the provided set, please add them. Otherwise, output the existing top-level topics as identified in the document.}

\texttt{[Top-level topics]}\\
\texttt{"[1] Topic A"}\\

\texttt{[Examples]}\\
\texttt{Example 1: Adding "[1] Topic B"}\\
\texttt{Document:}\\
\texttt{Topic B Document}\\

\texttt{Your response:}\\
\texttt{[1] Topic B: Definition}\\

\texttt{Example 2: Duplicate "[1] Topic A", returning the existing topic}\\
\texttt{Document:}\\
\texttt{Topic A Document}\\

\texttt{Your response:}\\
\texttt{[1] Topic A: Definition}\\

\texttt{[Instructions]}\\
\texttt{Step 1: Determine topics mentioned in the document.}\\
\texttt{- The topic labels must be as generalizable as possible.}\\
\texttt{- The topics must reflect a SINGLE topic instead of a combination of topics.}\\
\texttt{- The new topics must have a level number, a short general label, and a topic description.}\\
\texttt{- The topics must be broad enough to accommodate future subtopics.}\\
\texttt{- The final topic list must provide comprehensive topic coverage over the entire document. Output as many topics as needed to accomplish this instruction}\\
\texttt{Step 2: Perform ONE of the following operations:}\\
\texttt{1. If there are already duplicates or relevant topics in the hierarchy, output those topics and stop here.}\\
\texttt{2. If the document contains no topic, return "None".}\\
\texttt{3. Otherwise, add your topic as a top-level topic. Stop here and output the added topic(s). DO NOT add any additional levels.}\\

\texttt{[Document]}\\
\texttt{\{DOCUMENT\}}\\

\texttt{Please ONLY return the relevant or modified topics at the top level in the hierarchy.}\\
\texttt{[Your response]}
\end{tcolorbox}
\label{sec:appendix2}

\section{Topics Merging Prompt}
\label{sec:appendix3}

\begin{tcolorbox}
\footnotesize
\texttt{You will receive a list of topics that belong to the same level of a topic hierarchy. Your task is to merge topics that are paraphrases or near duplicates of one another. Return "None" if no modification is needed.}

\texttt{[Examples]}\\
\texttt{Example 1: Merging topics ("[1] Employer Taxes" and "[1] Employment Tax Reporting" into "[1] Employment Taxes")}\\
\texttt{Topic List:}\\
\texttt{[1] Employer Taxes: Mentions taxation policy for employer}\\
\texttt{[1] Employment Tax Reporting: Mentions reporting requirements for employer}\\
\texttt{[1] Immigration: Mentions policies and laws on the immigration process}\\
\texttt{[1] Voting: Mentions rules and regulation for the voting process}\\
\texttt{Your response:}\\
\texttt{[1] Employment Taxes: Mentions taxation report and requirement for employer ([1] Employer Taxes, [1] Employment Tax Reporting)}\\

\texttt{Example 2: Merging topics ("[2] Digital Literacy" and "[2] Telecommunications" into "[2] Technology")}\\
\texttt{Topic List:}\\
\texttt{[2] Mathematics: Discuss mathematical concepts, figures and breakthroughs.}\\
\texttt{[2] Digital Literacy: Discuss the ability to use technology to find, evaluate, create, and communicate information.}\\
\texttt{[2] Telecommunications: Mentions policies and regulations related to the telecommunications industry, including wireless service providers and consumer rights.}\\
\texttt{Your response:}\\
\texttt{[2] Technology: Discuss technology and its impact on society. ([2] Digital Literacy, [2] Telecommunications)}\\

\texttt{[Rules]}\\
\texttt{- Perform the following operations as many times as needed:}\\
\texttt{- Merge relevant topics into a single topic.}\\
\texttt{- Do nothing and return "None" if no modification is needed.}\\
\texttt{- When merging, the output format should contain a level indicator, the updated label and description, followed by the original topics.}\\

\texttt{[Topic List]}\\
\texttt{\{topic list\}}\\

\texttt{Output the modification or "None" where appropriate. Do not output anything else.}\\
\texttt{[Your response]}
\end{tcolorbox}

\section{Topic Assignment Prompt}
\label{sec:appendix5}

\begin{tcolorbox}
\footnotesize
\texttt{You will receive a document and a topic list. Assign the document to the most relevant topics. Then, output the topic labels, assignment reasoning and supporting quotes from the document. DO NOT make up new topics or quotes.}\\
\\
\texttt{Here is the topic list:}\\
\texttt{\{TOPIC LIST\}}\\

\texttt{[Instructions]}\\
\texttt{1. Topic labels must be present in the provided topic hierarchy. You MUST NOT make up new topics.}\\
\texttt{2. The quote must be taken from the document. You MUST NOT make up quotes.}\\
\texttt{3. If the assigned topic is not on the top level, you must also output the path from the top-level topic to the assigned topic.}\\

\texttt{[Document]}\\
\texttt{\{SINGLE PARAGRAPH\}}\\

\texttt{[Your response]}
\end{tcolorbox}

\section{Survey Questionnaire}
\label{sec:survey}

\noindent \textbf{Pre-Analysis Survey: } This survey had the following questions to study the annotators perception and prior experiences of using LLMs. The results are discussed in Table ~\ref{tab:survey_pre}.

\begin{tcolorbox}
    \footnotesize
    \texttt{1. Have you used LLM-based tools before? \\
    2. How much do you expect to trust the recommendations made by LLMs?\\
    3. How reliable do you expect the LLMs output to be?\\
    4. How accurate do you expect the LLM recommendations to be?\\
    5. How confusing do you expect the LLM recommendations to be?\\
    6. Do you think you will prefer completing tasks with or without the recommendations of an LLM?\\
    7. What are your initial expectations? Do you think having LLM suggestions will help with the analysis?\\
    8. What concerns do you have about using LLMs for your tasks?\\}
\end{tcolorbox}

\begin{table}[h]
\centering
\begin{tabularx}{\textwidth}{p{9cm}l}
\toprule
\textbf{Question} & \textbf{Response Distribution} \\
\midrule
Have you used LLM-based tools before? & 
Yes: 100\% \\
\midrule
How much do you expect to trust the recommendations  & 
1 - Not at all: 0\% \\
made by LLMs? & 2 - Slightly: 25\% \\
& 3 - Neutral: 50\% \\
& 4 - I trust to some extent: 25\% \\
& 5 - Completely: 0\% \\
\midrule
How reliable do you expect the LLMs output to be? & 
1 - Not reliable at all: 0\% \\
& 2 - Slightly reliable: 25\% \\
& 3 - Neutral: 50\% \\
& 4 - Reliable: 25\% \\
& 5 - Very reliable: 0\% \\
\midrule
How accurate do you expect the LLM recommendations  & 
1 - Not accurate at all: 0\% \\
to be? & 2 - Slightly accurate: 25\% \\
& 3 - Neutral: 25\% \\
& 4 - Accurate: 50\% \\
& 5 - Very accurate: 0\% \\
\midrule
How confusing do you expect the LLM recommendations  & 
1 - Not confusing at all: 25\% \\
to be? & 2 - Slightly confusing: 50\% \\
& 3 - Moderately confusing: 25\% \\
& 4 - Confusing: 0\% \\
& 5 - Very confusing: 0\% \\
\midrule
Do you think you will prefer completing tasks with or  & 
With LLM recommendations: 100\% \\
without the recommendations of an LLM? & Without LLM recommendations: 0\% \\
\bottomrule
\end{tabularx}
\caption{Pre-Analysis survey questions and corresponding response distributions for LLM-based tools.}
\label{tab:survey_pre}
\end{table}

\pagebreak

\noindent \textbf{Post-Analysis Survey: } This survey had the following questions to study the annotators experience and thoughts after completing the stage 2 of the study. The results of the Post-Analysis Survey are discussed in Table~\ref{tab:surv_p}.

\vspace{-950pt}

\begin{tcolorbox}
    \footnotesize
    \texttt{1. After seeing the LLM recommendations, do you prefer completing tasks with or without it? \\
    2. How much do you trustthe recommendations made by LLMs? \\   
    3. How reliable did you find the LLM's outputs? \\
    4. How confusing did you find the LLM's recommendations? \\
    5. How useful did you find the LLM in completing your tasks? \\
    6. How accurate were the LLM recommendations compared to your expectations?\\
    7. How would you rate the quality of the LLM's recommendations?\\
    8. How would you rate your overall experience of analysis with the LLM recommendations? \\
    9. Would you recommend using an LLM for similar tasks to others? \\
    10. How easy was it to integrate the LLM recommendations into your workflow? \\
    11. After seeing the recommendations, what concerns do you still/now have about using LLMs for your tasks? \\}
\end{tcolorbox}

\begin{table}
\centering
\begin{tabularx}{\textwidth}{p{9cm}l}
\toprule
\textbf{Question} & \textbf{Response Distribution} \\
\midrule
After seeing the LLM recommendations, do you prefer  & 
With LLM recommendations: 100\% \\
completing tasks with or without it? & Without LLM recommendations: 0\% \\
\midrule
How much do you trust the recommendations made  & 
1 - Not at all: 0\% \\
by LLMs? & 2 - Slightly: 25\% \\
& 3 - Neutral: 50\% \\
& 4 - I trust to some extent: 25\% \\
& 5 - Completely: 0\% \\
\midrule
How reliable did you find the LLM's outputs? & 
1 - Not reliable at all: 0\% \\
& 2 - Slightly reliable: 0\% \\
& 3 - Neutral: 75\% \\
& 4 - Reliable: 25\% \\
& 5 - Very reliable: 0\% \\
\midrule
How confusing did you find the LLM's recommendations? & 
1 - Not confusing at all: 0\% \\
& 2 - Slightly confusing: 50\% \\
& 3 - Moderately confusing: 50\% \\
& 4 - Confusing: 0\% \\
& 5 - Very confusing: 0\% \\
\midrule
How useful did you find the LLM in completing your tasks? & 
1 - Not useful at all: 0\% \\
& 2 - Slightly useful: 0\% \\
& 3 - Neutral: 25\% \\
& 4 - Useful: 50\% \\
& 5 - Very useful: 25\% \\
\midrule
How accurate were the LLM recommendations compared & 
1 - Not accurate at all: 0\% \\
to your expectations? & 2 - Slightly accurate: 0\% \\
& 3 - Neutral: 25\% \\
& 4 - Accurate: 50\% \\
& 5 - Very accurate: 25\% \\
\midrule
How would you rate the quality of the LLM's & 
1 - Very poor: 0\% \\
 recommendations? & 2 - Poor: 25\% \\
& 3 - Neutral: 50\% \\
& 4 - Good: 25\% \\
& 5 - Excellent: 0\% \\
\midrule
How would you rate your overall experience of analysis  & 
1 - Very poor: 0\% \\
with the LLM recommendations? & 2 - Poor: 0\% \\
& 3 - Neutral: 25\% \\
& 4 - Good: 75\% \\
& 5 - Excellent: 0\% \\
\midrule
Would you recommend using an LLM for similar tasks  & 
Yes: 75\% \\
to others? & Maybe: 25\% \\
& No: 0\% \\
\midrule
How easy was it to integrate the LLM recommendations into your workflow? & 
Very easy: 100\% \\
\bottomrule
\end{tabularx}
\caption{Post-Analysis survey questions and corresponding response distributions for LLM-based tools.}
\label{tab:surv_p}
\end{table}
\label{tab:survey_post}

\pagebreak

\section{Hyperparameter Tuning}
\label{sec:append_hyperparameter}

We tested many different temperatures when calling the model through API. We settled on a temperature of 0.2, as it provides a low degree of randomness, while also producing descriptive topics and definitions suitable for annotator interaction. 

\vspace{35pt}

\section{Label Studio Interface with Mock Annotations}
\label{sec:append_interface}

\begin{figure*}[h]
    \centering
    \includegraphics[width=\linewidth, keepaspectratio]{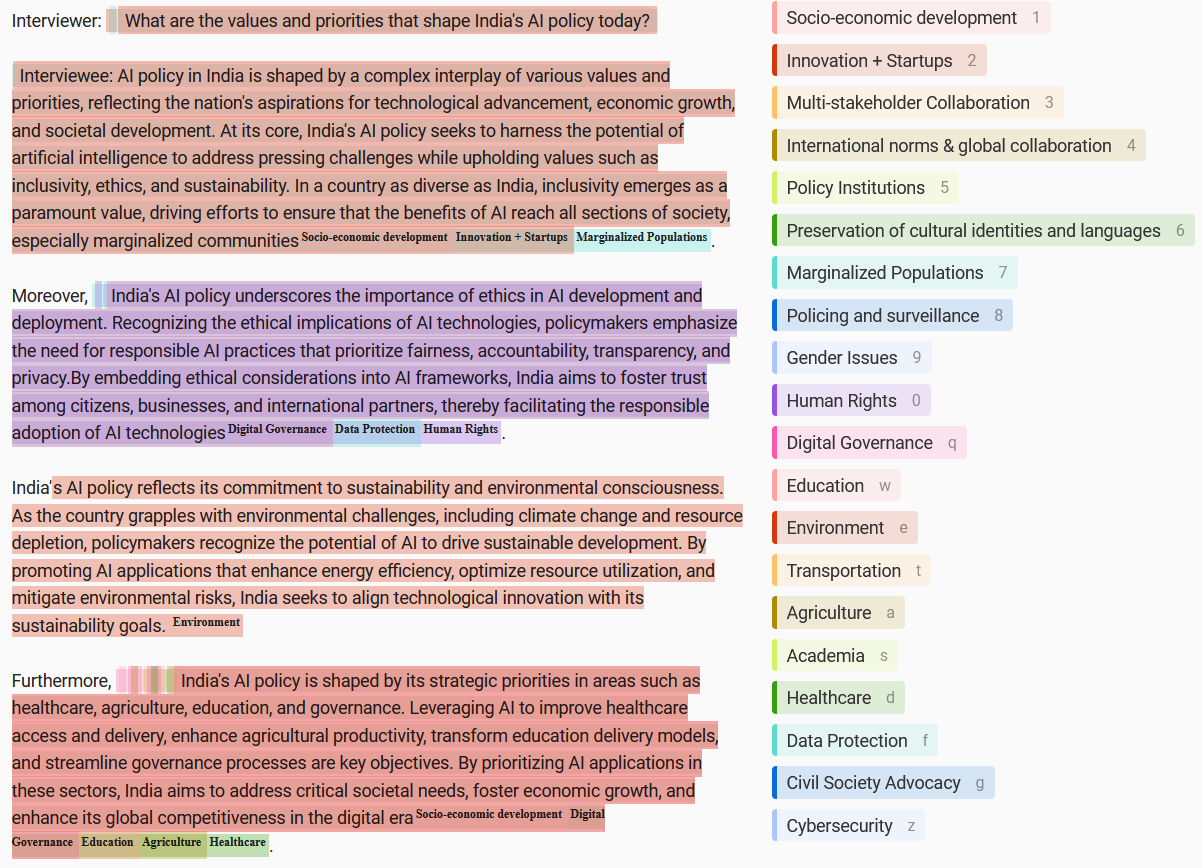}
    \caption{An example of the Label Studio GUI using a mock interview. In order to protect interviewee anonymity, interviews will not be released.}
    \label{fig:label_studio}
\end{figure*}

\vspace{35pt}

\section{Stage 1: Topic Lists}
From Stage 1 we compiled two topic lists. They are discussed is Tables~\ref{tab:final_topic_list} and \ref{tab:llm_topic_list}.
\label{sec:appendix6}

\begin{table*}
\centering
\begin{tabular}{p{5cm}p{10cm}}
\toprule
\textbf{Label Name} & \textbf{Label Definition} \\
\midrule
Socio-economic development & Emphasis on development outcomes including decreasing income inequality, improving health systems and access to health, and higher standards of living. Economic growth. \\
Innovation and Startups & Startups are emphasized as an important stakeholder and innovation emphasized as a key goal. \\
Multi-stakeholder Collaboration & Policies, programs, and dialogues between government, industry, and civil society groups including academia (triple-helix relationships). Includes public-private partnerships. \\
International norms \& global collaboration & Matters related to how the international community and their norms/regulations might have impacted regulations and policy in this case. (for ex: GDPR) \\
Policy Institutions & What institution is involved with developing, implementing and executing policy and regulations. Includes regulatory bodies, think-tanks… \\
Marginalized Populations & Groups of people who experience discrimination and exclusion due to unequal power relationships across social, political, economic, and cultural dimensions. \\
Policing and Surveillance & Elements of policy which use AI and technical tools for the purpose of policing and surveilling citizens. Also elements of concern over tools being used for policing and the surveillance of citizens. \\
Gender Issues & This includes examining gender inequality, roles, and biases in various societal contexts. \\
Human Rights & Matters pertaining to the protection or the degradation/non-protection of HRs. Matters related to how technology and AI might result in declines in citizen freedom. \\
Digital Governance & The use of digital technologies and practices by governments to enhance the access and delivery of government services to benefit citizens, businesses, and other stakeholders. This includes the implementation of digital tools, platforms, and policies to improve government operations, engage citizens, and foster transparency. \\
Education & Promotion and regulation of the confluence of AI and the education sector. \\
Environment & Promotion and regulation of the confluence of AI and the environmental sector. \\
Transportation & Promotion and regulation of the confluence of AI and the transportation sector. \\
Agriculture & Promotion and regulation of the confluence of AI and the agriculture sector. \\
Academia & Promotion and regulation of the confluence of AI and the academia sector. \\
Healthcare & Promotion and regulation of the confluence of AI and the healthcare sector. \\
Data Protection & Norms and specific policies related to the protection of citizen data online. \\
Civil Society Advocacy & How involved is civil society in dialoguing with the policy process and giving their perspective to shape things. \\
Cybersecurity & Concerns and regulations to deal with online fraud and criminal activity that exploits citizen data and ease of contacting citizens. \\
Preservation of cultural identities and languages & Preservation of cultural identity and languages of marginalized groups. \\
\bottomrule
\end{tabular}
\caption{Stage 1 Final Topic List curated by Annotators}
\label{tab:final_topic_list}
\end{table*}

\onecolumn
\begin{longtable}{p{5cm}p{10cm}}
\toprule
\textbf{Label Name} & \textbf{Label Definition} \\
\midrule
\endfirsthead
\toprule
\textbf{Label Name} & \textbf{Label Definition} \\
\midrule
\endhead
\bottomrule
\caption{Stage 1 Topic List generated by the LLM} \\
\endfoot
Cybersecurity and Data Protection & The protection of internet-connected systems, including hardware, software, and data, from cyber threats, and the process of safeguarding important information from corruption, compromise, or loss. This area covers efforts to safeguard data and systems from unauthorized access, attacks, or damage, and involves the establishment of policies and regulations that protect personal and organizational data from unauthorized access, use, disclosure, disruption, modification, or destruction. \\
Digital Governance & The use of digital technologies and practices by governments to enhance the access and delivery of government services to benefit citizens, businesses, and other stakeholders. This includes the implementation of digital tools, platforms, and policies to improve government operations, engage citizens, and foster transparency. \\
Artificial Intelligence (AI) and Ethics & The study and development of AI technologies that consider ethical principles and values. This involves addressing the moral implications and societal impacts of AI, including issues of fairness, accountability, transparency, and the protection of human rights in the design, development, and deployment of AI systems. \\
Economic Development through Digitization & The process of leveraging digital technologies to drive economic growth, innovation, and improved standards of living. This includes the transformation of traditional economies into digital economies, where digital information and technologies play a central role in economic activities, creating new opportunities for businesses and societies. \\
Startup Ecosystem Development & Focuses on the support and growth of startups through policies, incubation programs, and partnerships. This includes fostering innovation, providing resources for startups, and creating an environment conducive to entrepreneurial success. \\
Education Enhancement and Innovation & Focuses on the integration of technology in education to improve learning outcomes, access to education, and the development of digital skills, and encourages the development of a problem-solving mindset from a young age through initiatives like tinkering labs in schools. This topic covers the integration of advanced technologies into education to foster innovation and creativity among students. \\
Global Collaboration & Highlights the importance of international partnerships and knowledge exchange to drive innovation, address global challenges, and foster economic growth. This includes collaborations at various levels, from schools to industries, to leverage technology and innovation for societal benefit. \\
Socio-Economic Development & Focuses on leveraging innovation and technology to address socio-economic challenges, including poverty, education, healthcare, and infrastructure. This involves creating opportunities for job creation, economic growth, and improving the quality of life in underserved communities. \\
Digital Transformation and Infrastructure & Emphasizes the role of digital technologies in transforming societies and economies. This includes the development of digital infrastructure to support innovation, such as mobile technology, internet access, and digital payment systems, to ensure inclusivity and accessibility for all. \\
Sustainable Development and SDGs Alignment & Encourages innovations that align with the Sustainable Development Goals (SDGs) to ensure that technological advancements contribute positively to environmental sustainability, social equity, and economic viability. This includes fostering a culture of innovation that considers the impact on the planet and society. \\
Marginalized Populations & Groups of people who experience discrimination and exclusion due to unequal power relationships across social, political, economic, and cultural dimensions. \\
Language and Linguistics & The study and analysis of the structure, development, and usage of languages, including their sociopolitical and cultural impacts. \\
Gender Studies & An interdisciplinary field exploring gender identity, expression, and gendered representation as central categories of analysis; this includes examining gender inequality, roles, and biases in various societal contexts. \\
Education and Literacy & The exploration of teaching and learning processes, literacy development, and educational systems. This includes access to education, pedagogical strategies, and the role of language and technology in education. \\
Cultural Identity and Preservation & The study of how cultures and communities maintain, preserve, and transform their identities, practices, and languages in the face of globalization, technological change, and sociopolitical pressures. \\
Technology Governance & Involves the policies, frameworks, and standards that guide the development, deployment, and management of technology within societies. It aims to ensure that technology serves the public good, addresses ethical considerations, and mitigates potential harms. \\
Agriculture and Food Security & Focuses on the application of technology and innovative practices to improve agricultural productivity, food security, and sustainability. This includes advancements in crop management, pest control, and the use of AI and drones for agricultural improvement. \\
Public-Private Partnerships & Highlights the collaboration between the public sector, private industry, and civil society to foster innovation, address societal challenges, and drive economic growth through technology. \\
Data Governance and Privacy & Addresses the management, sharing, and protection of data in the digital age. This includes the development of policies and frameworks to ensure data privacy, security, and ethical use of data. \\
Health Innovation & Encompasses the development and application of new technologies and approaches to improve health outcomes. This includes the use of AI for early disease detection, digital health advisories, and innovations in healthcare delivery. \\
Urban Transformation & Involves the use of technology to address urban challenges and improve city living. This includes smart city initiatives, urban planning technologies, and solutions for sustainable urban development. \\
Circular Economy and Sustainability & Concentrates on the development of systems and technologies that promote resource efficiency, waste reduction, and the sustainable management of natural resources. This includes initiatives in plastic recycling and the promotion of circular economic models. \\
\label{tab:llm_topic_list} 
\end{longtable}

\end{document}